# Semi-Supervised Learning with Multiple Imputations on Non-Random Missing Labels

Jason Lu[1+*], Michael Ma[2+], Huaze Xu[3+], Zixi Xu[4+]

[1]Millburn High School, Millburn, 07078, USA
[2]Franklin High School, Elk Grove, 95758, USA
[3]Beijing University of Posts and Telecommunication, Beijing, 100080, China
[4]Yew Wah International Education School of Guangzhou, Guangzhou,510000, China
* Corresponding author. Email: jasonlu05j@gmail.com
+These authors contributed equally to this work and should be considered co-first authors.

## ABSTRACT

Semi-Supervised Learning (SSL) is implemented when algorithms are trained on both labeled and unlabeled data. This is a very common application of ML as it is unrealistic to obtain a fully labeled dataset. Researchers have tackled three main issues: missing at random (MAR), missing completely at random (MCAR), and missing not at random (MNAR). The MNAR problem is the most challenging of the three as one cannot safely assume that all class distributions are equal. Existing methods, including Class-Aware Imputation (CAI) and Class-Aware Propensity (CAP), mostly overlook the non-randomness in the unlabeled data. This paper proposes two new methods of combining multiple imputation models to achieve higher accuracy and less bias. 1) We use multiple imputation models, create confidence intervals, and apply a threshold to ignore pseudo-labels with low confidence. 2) Our new method, SSL with De-biased Imputations (SSL-DI), aims to reduce bias by filtering out inaccurate data and finding a subset that is accurate and reliable. This subset of the larger dataset could be imputed into another SSL model, which will be less biased. The proposed models have been shown to be effective in both MCAR and MNAR situations, and experimental results show that our methodology outperforms existing methods in terms of classification accuracy and reducing bias.

**Keywords:** Semi-supervised learning, multiple imputations, fixmatch, pseudo-label, Missing Not At Random

## 1. INTRODUCTION

As we have already established, SSL leverages unlabeled data when labeled data is scarce. In many cases, it is impractical to obtain a fully labeled dataset, as it can be extremely expensive and time-consuming. Therefore, applying ML models to partially labeled datasets is a common problem that many researchers have attempted to address. Many contemporary SSL methods train a model based on the labeled data, then use that model to generate labels for the unlabeled data. This process is called pseudo-label prediction, and is used in CAI [1-4]. CAI is extremely effective in MAR and MCAR scenarios, as imputation can accurately predict the unlabeled data's true values with minimal bias. However, in the MNAR scenario, this is not the case as the data is not randomly unlabeled. There are often highly imbalanced class distributions, and specific reasons for the data being unlabeled. It is impossible to get to the root of the problem, but assuming the labeled and unlabeled data are evenly distributed would be a fatal error. There are other prevailing methods include Class-Aware-Propensity (CAP), which uses propensity scores to reweight missing data points based on class information and distribution. However, like with CAI, these methods are quite effective in both the MAR and MCAR scenarios, but fall short when handling MNAR. [5, 6] The non-randomness of the class distribution throws a wrench in common SSL methodologies. Thus, a combination of the two methods, or Class-Aware Doubly Robust (CADR) model, has also been proposed. CADR uses both CAI and CAP together, and the ICLR 22 Paper demonstrated that as long as one method achieves accurate results, the CADR model can be used. This is much more efficient and effective than implementing only one algorithm.

However, these methods still have their limitations. This paper proposes a new method that combines multiple imputation models to increase the amount of reliable labeled data and reduce bias in SSL models. Specifically, we propose using multiple imputation models to generate confidence intervals for the missing data, and then setting a threshold for accepting imputations with high confidence. This approach increases the amount of labeled data available for training, and potentially reduces the bias introduced by the imputation models. Moreover, we also propose another method, called SSL with De-biased Imputations (SSL-DI), that further reduces bias by selecting a subset of reliable data from the imputed data based on the agreement of multiple SSL models. We first train multiple SSL models on the imputed data, and then select the subset of data where the models predict similarly. This subset of data is considered more reliable than the rest of the imputed data and is then used to train another SSL model, such as FixMatch, which is less biased [4].

We perform experiments on both MAR and MNAR scenarios to determine the effectiveness of our proposed methods. The experimental results demonstrate the improvements of our approach, as it achieves better classification accuracy and less bias compared to existing methods, such as CAI, CAP, and CADR [5]. Additionally, we show that varying the number of imputation models, the threshold for accepting imputations, the number of SSL models, and the epsilon used for determining the similarity between models can have different effects on the results, suggesting that our method is versatile and can be adapted to various scenarios.

The key contributions of this work are summarized as follows:

1) We analyze the challenging problem of the Missing Not At Random (MNAR) setting in semi-supervised learning, as well as the shortcomings of existing SSL methods that attempt to combat it.

2) We propose two new methods using multiple SSL models. The first is to construct multiple imputation models and use confidence level to choose more accurate prediction labels.

3) Our second proposal is called SSL with De-biased Imputations (SSL-DI), and it aims to find a more reliable subset of the original dataset to train another SSL model on.

4) Our proposed methods achieve competitive performances in both MNAR and MAR settings

## 2 RELATED WORKS

### 2.1 Ensemble learning

It aims to combine multiple separate models to achieve better generalization performance. Co-training designed to use two learning algorithm that trained by two different views when only a small set of labeled data are available[6]. Similar approaches are used in many areas of machine learning: Co-teaching deep learning with noisy labels, they train two neural network models at the same time and decides what data to be use for training and Co-teaching+ train two networks, both networks feed forward and predict all data, retaining data with inconsistent predicted results, in which each network selects its own small loss data, but backpropagates the small loss data from the peer network and updates its own parameters[7,8]. Deep Co-training it designed to train many deep neural networks into different views and encourage view difference by using adversarial examples. This prevents each model from ending up the same [9].

### 2.2 Long-tail distribution (MNAR)

An imbalanced or long-tailed label distribution in the real-word classification problems, for example products on a website that sell well usually get more feedback from customers, and products that don't sell well get less. Several articles have talked about how to train an unbiased model under this situation. ABC, an auxiliary balance classifier with extra regularization terms reduces the deviation of learning model caused by imbalance in class distribution [10]. CAPL by exploited unlabeled data and dynamically adjusting the threshold for selecting pseudo labels. Given more accurate pseudo labels, the model will have better performance under such imbalance data distribution. CADR by combining the CAP score (An improved classifier trained using the unlabeled data in the biased labeled data) and CAI (encourage the rare class training. Decrease or increase the threshold for rare class dynamically and generate the pseudo label) to train an unbiased SSL model [1,2].

### 2.3 MCAR

where both labeled and unlabeled data share the same class distribution. In this case, semi-supervised learning model can make good use of unlabeled data to train a model. Many articles have also mentioned how to train the model in this case. FixMatch generate the pseudo labels and only keep the pseudo labels with high confidence [11]. MixMatch, It can guess the low entropy labels of the unlabeled examples of data enhancement and use MixUp to mix the labeled and unlabeled data [12]. ReMixMatch, an improved MixMatch, by Distribution alignment (Encourage the marginal distribution of predictions for unlabeled data to approach the marginal distribution of the ground-truth labels.) and Augmentation anchoring (Provide multiple strongly augmented versions of the input into the model and encourage prediction of weakly augmented versions with each output close to the same input.) [13].

## 3. PROBLEM SETUP

Dataset D is divided into two subsets $D_u$ and $D_l$. The former is a dataset that does not contain labeled data, while the latter contains complete observed data. To be specific, $D_u$ is denoted as $D_u = \{x_i\}_{i=1}^{U}$ whereas $D_l$ is defined as $D_l = \{(x^{(i)}, y^{(i)})\}_{i=1}^{L}$. Notice that under most circumstances, the size of $D_u$ (U) is much larger than that of $D_l$ (L).

The goal is to learn from the datasets and obtain a function $f: X \rightarrow Y$ that performs well on the entire input space. In the CADR learning algorithm, a new dimension M is introduced to D to indicate whether the label is missing. Thus, D is represented as $D = (X, Y, R) = \{(x^{(i)}, y^{(i)}, r^{(i)})\}_{i=0}^{N}$, where $m \in [0, 1]$: m=1 indicates that y is unobserved, and m=0 indicates that y has been observed. The size of the set N=U+L, where U is the size of the unlabeled dataset $D_u$ and L is the size of labeled dataset $D_l$. Unlike MCAR (missing completely at random) and MCAR (missing completely at random), MNAR means that the missing label itself contains information. ($i.e. Y$ ( $P(Y|M, X) \neq P(Y|X)$). The aim is to train a model based on D that can accurately predict the classification result of inputs in MNAR situation.

In CADR, learning is represented as $\hat{\theta}_{CADR} = arg \min_{\theta} l_{CADR} = arg \min_{\theta} l_{CAP} + l_{CAI} + l_{support}$

$$l_{CAP} = \frac{1}{N}\sum_{i=1}^{n} \frac{(1-r^{(i)})l_s(x^{(i)}, y^{(i)})}{p^{(i)}} \qquad (1)$$

$$l_{CAI} = \frac{1}{N}\sum_{i=1}^{n} (r^{(i)} l_u(x^{(i)}, q^{(i)}) I(con(q^{(i)}) > \tau(x^{(i)})) + (1 - r^{(i)}) l_s(x^{(i)}, y^{(i)}) \qquad (2)$$

$p^{(i)}$ is the propensity score, $l_s$ is the cross-entropy loss, $l_u$ is the L2 loss, $con$ is the confidence function, and $\tau$ is the threshold function.

$l_{CAP}$ is the loss measuring the discrepancy between the predicted and true labels, and is weighted by the propensity score to correct for any sampling bias. $l_{CAI}$ is the loss estimating the labels of the unlabeled samples and incorporates them into the learning process. It is weighted by the confidence function to ensure that only highly confident predictions are included.

CADR learning has been demonstrated to achieve better performance on multiple datasets compared to many other semi-supervised learning algorithms. Our aim is to improve upon CADR learning.

In this paper, we introduce two new methods to improve existing SSL algorithms in the MNAR data scenario.

## 4. METHODOLOGY

### 4.1. Our Method Integrating Two Fixmatch Model

(1) When two models are given labeled data, we use different orders and augmentations to give them different initial states.

(2) We maximize the use of information contained in both models to generate as many pseudo-labels as possible through two methods:

1. If either model's prediction confidence exceeds a threshold $\tau = 95\%$, the prediction is considered reliable.

2. If both models' prediction confidences exceed a lower threshold $\tau2$ ($\tau2 < \tau$) and the predictions are the same, the agreement is considered reliable. Lowering $\tau$ can generate more pseudo-labels, but it may reduce their accuracy. However, if stricter requirements are needed (i.e., the predictions of both models must be the same), this reduction in accuracy can be compensated for.

In addition, there are occasional special cases where (1) both models' prediction confidences exceed 95%, but their predictions differ, or (2) the prediction with a confidence over 95% differs from the agreement with a confidence exceeding τ2. In the first case, we randomly select one prediction as the pseudo-label. In the second case, we choose the prediction with a confidence over 95% as the pseudo-label. We demonstrate the validity of our approach to these special cases in ablation experiments.

Experimental results show that our method significantly increases the number of pseudo-labels obtained. Moreover, the accuracy of the pseudo-labels is even slightly improved compared to using only one FixMatch model.

(3) Traditional algorithms often improve the diversity between models by dividing samples into different subsets for each model. However, we found that dividing reliable pseudo-labels into different subsets for the two models is not a good approach. This only slightly improves the diversity between models, but in semi-supervised learning, supervised information is very valuable, so this approach wastes a lot of valuable information and is not worth the cost. Therefore, we allow both models to use all reliable pseudo-labels, but we give higher weights to simpler pseudo-labels for the model that finds them easier, so that the pseudo-labels provided by the other model have relatively lower weights. This can alleviate the convergence between the two models.

### 4.2. pseudo code

---

**Algorithm 1. Our algorithm with two models**

---

**Input:** Labeled batch for model 1 $X_1 = \{(x_{1b}, p_{1b}) : b \in (1, \ldots, B)\}$, Labeled batch for model 2 $X_2 = \{(x_{2b}, p_{2b}) : b \in (1, \ldots, B)\}$, unlabeled batch $U = \{u_b : b \in (1, \ldots, \mu B)\}$, confidence threshold $\tau$, lower confidence threshold $\tau_2$ unlabeled data ratio $\mu$, unlabeled loss weight $\lambda$, pseudo-label weight $w$.

$l_{s1} = \frac{1}{B} \sum_{b=1}^{B} H(p_{1b}, \alpha_1(x_{1b}))$ {Cross-entropy loss for labeled data for model 1}

$l_{s2} = \frac{1}{B} \sum_{b=1}^{B} H(p_{2b}, \alpha_2(x_{12b}))$ {Cross-entropy loss for labeled data for model 2}

**for** $b = 1$ **to** $\mu B$ **do**

    $q_{1b} = p_m(y|\alpha(u_b); \theta_1)$ {Compute prediction of $u_b$ by model 1}

    $q_{2b} = p_m(y|\alpha(u_b); \theta_2)$ {Compute prediction of $u_b$ by model 2}

    **if** $max(q_{1b} > \tau)$

        $m_{1b} = w$ {It is a reliable *pseudo-label according to model 1*} (The 'm' of 'm1b' means 'mask')

    **elif** $max(q_{2b}) > \tau$ or $(max(q_{1b}) > \tau_2$ and $\tau$ $max(q_{2b}) > \tau_2$ and $arg\,max\,(q_{1b}) = arg\,max(q_{2b}))$

        $m_{1b} = 1$ {It is a reliable pseudo-label according to model 2}

    **else**

        $m_{1b} = 0$ {It is not a reliable pseudo-label}

    **if** $max(q_{2b} > \tau)$

        $m_{2b} = w$ {It is a reliable pseudo-label according to model 2}

    **elif** $max(q_{1b}) > \tau$ or $(max(q_{1b}) > \tau_2$ and $\tau$ $max(q_{2b}) > \tau_2$ and $arg\,max\,(q_{1b}) = arg\,max(q_{2b}))$

        $m_{2b} = 1$ {It is a reliable pseudo-label according to model 1}

    **else**

$$m_{2b} = 0 \ \{\textit{It is not a reliable pseudo-label}\}$$

**if** $max(q_{mb}) > \tau$ {*There is a 50% chance that m is 1 and a 50% chance that m is 2. When m is 1 n is 1. When m is 2 n is 1*}

$$r_b = argmax(q_{mb}) \quad (\textit{The 'r' of 'rb' means 'reliable pseudo-label'})$$

**elif** $max(q_{nb}) > \tau$

$$r_b = argmax(q_{nb})$$

**elif** $max(q_{1b}) > \tau_2$ and $max(q_{2b}) > \tau_2$ and $arg\,max(q_{1b}) = arg\,max(q_{2b})$

$$r_b = argmax(q_{1b})$$

**else**

$$r_b = 0 \ \{\textit{randomly pick a pseudo-label}\}$$

**end for**

$$l_{u1} = \frac{1}{\mu B} \sum_{b=1}^{\mu B} H(r_b, p_m(y|A(\mu_{1b}))) * m_{1b} \quad \{\textit{Cross-entropy loss for unlabeled data for model 1}\}$$

$$l_{u2} = \frac{1}{\mu B} \sum_{b=1}^{\mu B} H(r_b, p_m(y|A(\mu_{2b}))) * m_{2b} \quad \{\textit{Cross-entropy loss for unlabeled data for model 2}\}$$

**return** $l_1 = l_{s1} + \lambda l_{u1}, l_2 = l_{s1} + \lambda l_{u2}$

---

### 4.3. Further explanation

(1) For other FixMatch-based algorithms, our integration method can be achieved with slight modifications to the above steps.

(2) This integration method can not only be used to integrate two models, but also to integrate multiple models. In this case, for example, when integrating multiple FixMatch models:

1. If the prediction confidence of any model exceeds $\tau = 95\%$, the prediction result with a confidence level of over 95% is considered reliable.

2. If the prediction confidence of at least two models exceeds $\tau_2 (\tau_2 < \tau)$ and they reach a consensus, the prediction result with a consensus and a confidence level of over τ2 is considered reliable.

3. If the prediction confidence of at least three models exceeds $\tau_3 (\tau_3 < \tau_2)$ and they reach a consensus, the prediction result with a consensus and a confidence level of over τ3 is considered reliable

n. If the prediction confidence of at least n models exceeds $\tau_n (\tau_n < \tau_{n-1})$ and they reach a consensus, the prediction result with a consensus and a confidence level of over τn is considered reliable.

As with integrating two models, when there are conflicts between the above prediction methods, the method with the larger τ value is used as the main method. If there are conflicts within the same τ method, a reliable pseudo-label is randomly selected.

## 5. EXPERIMENT SETUP

### 5.1 Dataset

We conducted our experiments on the CIFAR10 Dataset [5]. It consists of 60,000 32x32 color images in 10 classes. The different classes represent images of airplanes, cars, birds, cats, deer, dogs, frogs, horses, ships, and trucks.

## 5.2 Code

Since the original FixMatch is too large and complex, it is difficult to train multiple models at the same time based on it [14]. Therefore, instead of using the official code of FixMatch, we built our code based on an unofficial code of FixMatch on GitHub.

## 5.3 Preparation

Although our idea is very simple in theory, there are still many unexpected details to consider when designing its code. To this end, we conduct a series of preliminary experiments to determine the best way to implement our ideas. The comparison of the accuracy of the various algorithms we designed with the original FixMatch algorithm (light blue curve) on the test set is shown in Figure 1, and the comparison of the loss on the test set is shown in Figure 2. It is worth noting that we observed that when changing some details of the algorithm, the effect on the accuracy of the algorithm on the test set is not continuous. Specifically, as the operation cycle increases, the accuracy curves of each algorithm always move closer to certain specific curves (see the longest three curves in Figure 1), which may be caused by the similarity of some deep mechanisms of. In addition, we also note that using: $l_1 = l_{x1} + \lambda(l_{u1} + l_{u1})$ is clearly better than using $l_1 = l_{x1} + \lambda l_{u1}$, which is quite surprising. We will try to explain this phenomenon in our future work.

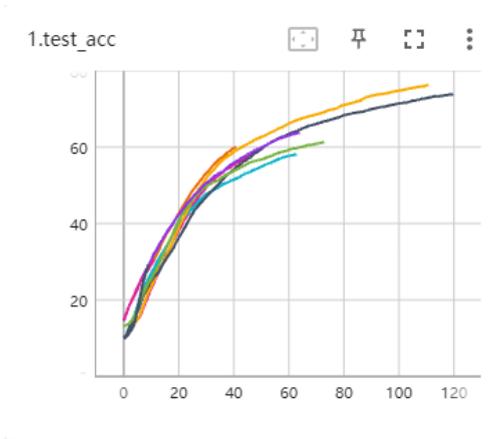
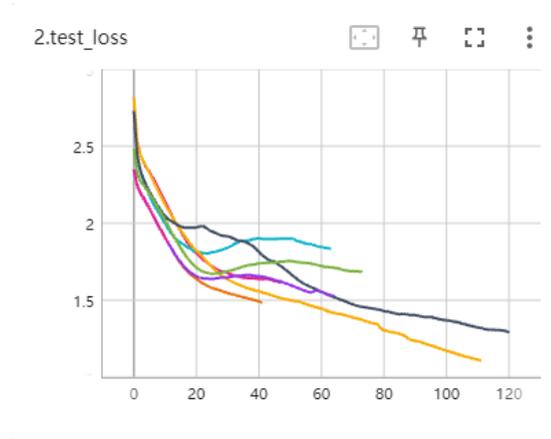

Figure 1. Comparison of Accuracy      Figure 2. Comparison of Loss

Note：（Light blue curve shows the accuracy of our algorithm）

## 6. EXPERIMENTAL RESULTS AND ANALYSIS

Due to time constraints, we put our two ideas into the same code and compared it with the original FixMatch. All the common hyperparameters of the two adopt the hyperparameters used for the cifar10 dataset in the FixMatch official paper [14]. According to the Ablation Studies on hyperparameters in the official FixMatch paper, we set the newly added hyperparameters in our code to 0.75. And, in order to increase the difficulty of the classification task, so as to better prove the good performance of our method under extreme conditions, we reduced the total parameters of the neural network from the default 1.47M to 0.3M, and randomly selected 150 examples as labeled examples (the number of samples per class is not necessarily 15). Our experiments are run on RTX3080 GPU.

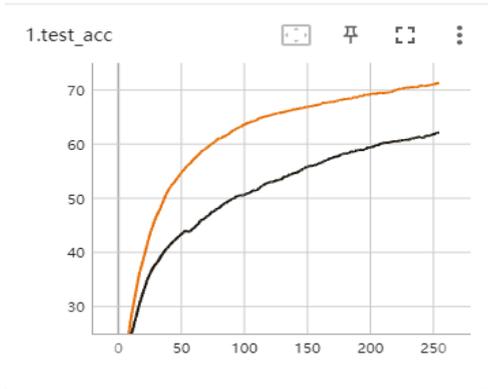

Figure 3. Comparison of accuracy

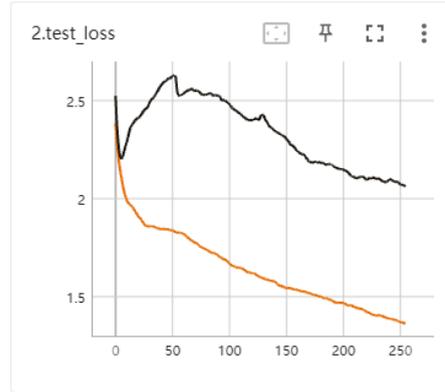

Figure 4. Comparison of loss on test set

Note：(Orange curve represents the trend of our algorithm while back represents Fixmatch)

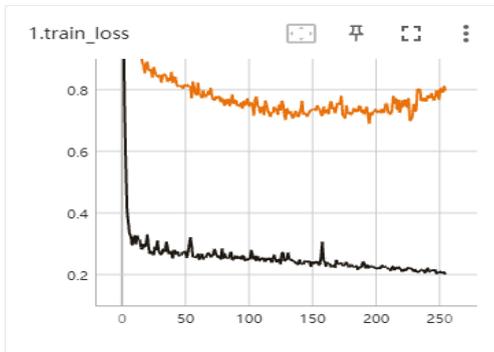

Figure 5. Comparison of loss on Training set

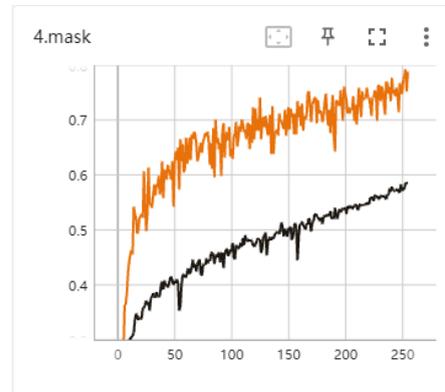

Figure 6. Trend of mask

In the above four figures, the black curve represents the original FixMatch, and the orange curve represents our method. Figure 3 represents the accuracy of the two methods on the test set. Although the two methods cannot be converged due to time constraints, it can be seen that our method can steadily improving accuracy by about 10% to 13%. Figure 4 represents the loss of the two methods on the test set. It can be seen that the loss of the original FixMatch on the test set has large fluctuations, and our method avoids this fluctuation. In Figure 5, the loss of the two methods on the training set is quite different. This is because the methods of calculating the loss on the training set of these two methods are quite different. Figure 6 represents for each train batch, what proportion of pseudo-labels are considered "credible" by these two methods. It can be seen that our method can steadily increase the proportion of accurate pseudo-labels by about 20%.

## 7. CONCLUSIONS

All in all, by combining our two ideas, we were able to steadily improve the performance of the original FixMatch under extreme conditions. Compared with other work on improving the performance of FixMatch under extreme conditions [1, 2], our improvement in accuracy is not very large. However, our method aims to give as many unlabeled examples as accurate as possible pseudo-labels in each batch of training when training multiple models capable of imputation at the same time. Since most of the current work on improving FixMatch does not involve this aspect of improvement [15, 16], our method has good compatibility. In addition, our method can be implemented with very little modification, so any semi-supervised learning method involving input can be upgraded by our method. In the next work, we will examine the performance of our method on other data sets, apply our method on various semi-supervised learning methods, and tune hyperparameters.

# REFERENCES

[1] Gui, Qian, Xinting Wu, and Baoning Niu. "Class-Aware Pseudo Labeling for Non-random Missing Labels in Semi-supervised Learning." 2022 IEEE Eighth International Conference on Multimedia Big Data (BigMM). IEEE, 2022.

[2] Hu, Xinting, et al. "On non-random missing labels in semi-supervised learning." arXiv preprint arXiv:2206.14923 (2022).

[3] Berthelot, David, et al. "Mixmatch: A holistic approach to semi-supervised learning." Advances in neural information processing systems 32 (2019).

[4] Kihyuk Sohn, David Berthelot, Chun-Liang Li, Zizhao Zhang, Nicholas Carlini, Ekin D. Cubuk, Alex Kurakin, Han Zhang and Colin Raffel. "Unofficial PyTorch implementation of "FixMatch: Simplifying Semi-Supervised Learning with Consistency and Confidence"" (2020).

[5] Krizhevsky, Alex, and Geoffrey Hinton. "Learning multiple layers of features from tiny images." (2009): 7.

[6] Blum, Avrim, and Tom Mitchell. "Combining labeled and unlabeled data with co-training." Proceedings of the eleventh annual conference on Computational learning theory. 1998.

[7] Han, Bo, et al. "Co-teaching: Robust training of deep neural networks with extremely noisy labels." Advances in neural information processing systems 31 (2018).

[8] Yu, Xingrui, et al. "How does disagreement help generalization against label corruption?." International Conference on Machine Learning. PMLR, 2019.

[9] Qiao, Siyuan, et al. "Deep co-training for semi-supervised image recognition." Proceedings of the European conference on computer vision (eccv). 2018.

[10] Lee, Hyuck, Seungjae Shin, and Heeyoung Kim. "Abc: Auxiliary balanced classifier for class-imbalanced semi-supervised learning." Advances in Neural Information Processing Systems 34 (2021): 7082-7094

[11] Sohn, Kihyuk, et al. "Fixmatch: Simplifying semi-supervised learning with consistency and confidence." Advances in neural information processing systems 33 (2020): 596-608.

[12] Berthelot, David, et al. "Mixmatch: A holistic approach to semi-supervised learning." Advances in neural information processing systems 32 (2019).

[13] Berthelot, David, et al. "Remixmatch: Semi-supervised learning with distribution alignment and augmentation anchoring." arXiv preprint arXiv:1911.09785 (2019).

[14] Kihyuk Sohn, David Berthelot, Chun-Liang Li, Zizhao Zhang, Nicholas Carlini, Ekin D. Cubuk, Alex Kurakin, Han Zhang and Colin Raffel. "A simple method to perform semi-supervised learning with limited data." (2020).

[15] Lukasiewicz, Thomas, and Jianfeng Wang. "NP− Match: When Neural Processes meet Semi− Supervised Learning." (2022).

[16] Zhang, Bowen, et al. "Flexmatch: Boosting semi-supervised learning with curriculum pseudo labeling." Advances in Neural Information Processing Systems 34 (2021): 18408-18419.